\documentclass[10pt,twocolumn,letterpaper]{article}

\usepackage{iccv}              % To produce the CAMERA-READY version
\definecolor{iccvblue}{rgb}{0.21,0.49,0.74}
\usepackage[pagebackref,breaklinks,colorlinks,allcolors=iccvblue]{hyperref}
\usepackage{multirow,colortbl}

\def\paperID{10372} % *** Enter the Paper ID here
\def\confName{ICCV}
\def\confYear{2025}

\title{HiTVideo: Hierarchical Tokenizers for Enhancing Text-to-Video Generation with Autoregressive Large Language Models}

\author{
    Ziqin Zhou\textsuperscript{1,2}\thanks{Work done during internship at Microsoft Research Asia.}\quad
    Yifan Yang\textsuperscript{2}\thanks{Corresponding authors.}\quad
    Yuqing Yang\textsuperscript{2}\quad
    Tianyu He\textsuperscript{2}\quad
    Houwen Peng\textsuperscript{2}\quad\\
    Kai Qiu\textsuperscript{2}\quad
    Qi Dai\textsuperscript{2}\quad
    Lili Qiu\textsuperscript{2}\quad
    Chong Luo\textsuperscript{2}\quad
    Lingqiao Liu\textsuperscript{1}\footnotemark[2]\\[8pt]
    \textsuperscript{1}The University of Adelaide\quad
    \textsuperscript{2}Microsoft Research Asia \\
    {\tt\small \{ziqin.zhou, lingqiao.liu\}@adelaide.edu.au, {yifanyang@microsoft.com}}
}

\begin{document}
\maketitle
\begin{abstract}
Text-to-video generation poses significant challenges due to the inherent complexity of video data, which spans both temporal and spatial dimensions. It introduces additional redundancy, abrupt variations, and a domain gap between language and vision tokens while generation. Addressing these challenges requires an effective video tokenizer that can efficiently encode video data while preserving essential semantic and spatiotemporal information, serving as a critical bridge between text and vision. Inspired by the observation in VQ-VAE-2 and workflows of traditional animation, we propose \textbf{HiTVideo} for text-to-video generation with hierarchical tokenizers. It utilizes a 3D causal VAE with a multi-layer discrete token framework, encoding video content into hierarchically structured codebooks. Higher layers capture semantic information with higher compression, while lower layers focus on fine-grained spatiotemporal details, striking a balance between compression efficiency and reconstruction quality. Our approach efficiently encodes longer video sequences (e.g., 8 seconds, 64 frames), reducing bits per pixel (bpp) by approximately 70\% compared to baseline tokenizers, while maintaining competitive reconstruction quality. We explore the trade-offs between compression and reconstruction, while emphasizing the advantages of high-compressed semantic tokens in text-to-video tasks. HiTVideo aims to address the potential limitations of existing video tokenizers in text-to-video generation tasks, striving for higher compression ratios and simplify LLMs modeling under language guidance, offering a scalable and promising framework for advancing text to video generation. \textbf{\textcolor{blue}{Demo page}}: \url{https://ziqinzhou66.github.io/project/HiTVideo}.
% \href{https://ziqinzhou66.github.io/project/HiTVideo}{\textcolor{blue}{Demo page}}

\end{abstract}    
\vspace{-2mm}
\section{Introduction}
\label{sec:intro}

Visual generation is a challenging and rapidly evolving task that has seen significant advancements over the years. Initially, GANs (Generative Adversarial Networks) \cite{goodfellow2014gan,karras2019stylegan} shows impressive capabilities in creating realistic images \cite{karras2017progressive,karras2019stylegan} and videos \cite{tulyakov2018mocogan,skorokhodov2022stylegan-v}. In recent years, the development of diffusion models \cite{rombach2022stablediffusion,dhariwal2021diffusion,ho2022imagen-v} and Large Language Models (LLMs) \cite{raffel2020t5,sanh2021flant5} has provided new solutions for tackling visual generation tasks, especially for generating content based on human instructions, such as text-to-video generation \cite{he2022lvdm,hong2022cogvideo}. Diffusion models have become a prominent approach in visual generation, particularly due to their ability to produce high-quality outputs by gradually refining noise through iterative denoising processes. On the other hand, LLMs have revolutionized unified multi-modal framework, enabling significant breakthroughs in natural language understanding and generative tasks. 

Despite these developments, training generative models for text-to-video generation presents significant challenges, particularly in visual encoding and multimodal alignment. Unlike image data, video sequences have greater redundancy and variability across frames, complicating the encoding process and making video tokenization especially challenging. An effective video tokenizer must strike a balance between reducing redundant information and preserving crucial spatiotemporal details to encode video content. Additionally, the gap between text tokens and visual tokens is a major barrier in aligning language and visual representations for generation tasks. A well-designed tokenizer can better represent semantic content, a critical requirement for generating coherent tokens cross language to visual. 

Related video autoencoder, such as MAGVIT-V2 \cite{yu2023magvitv2}, have demonstrated the utility of dynamic masking strategies in enhancing generation efficiency and reconstruction quality \cite{chang2022maskgit}. Similarly, VAR \cite{tian2024var} has shown that a coarse-to-fine strategy improves text-to-image generation. However, these prior methods primarily focus on static images or short sequences and do not directly address the unique challenges of autoregressive text-to-video generation.
% particularly for long video sequences.

To design a well video tokenizer, inspired by traditional animation workflows—where animators first create high-level semantic frameworks (e.g., storyboards) and then progressively add details—we propose a multi-layer codebook for video. Drawing on insights from VQ-VAE-2 \cite{razavi2019vqvae-2}, which demonstrates that high-level features effectively capture semantic information while low-level features excel at reconstructing details, our design integrates multi-layer tokenizers to address the challenges of text-to-video generation.

Recent works \cite{wu2024janus,wang2024emu3} utilize LLMs to unify visual and language tokens, facilitating effective alignment with language tokens in downstream tasks. Applying this paradigm to visual generation offers unique advantages. First, LLMs are highly effective at understanding and aligning complex semantics, making them particularly well-suited for tasks like text-to-video generation, where accurate semantic mapping between textual descriptions and video content is essential. Second, LLMs benefit from efficient cache computations and dependency management across time steps. Instead, they employ next token prediction \cite{yan2021videogpt} or masked prediction methods \cite{yu2023magvitv2} that are computationally efficient, especially with techniques like key-value caching (KV-cache)  \cite{luohe2024keep,chung2024scaling} that optimize resource usage. Moreover, the inherent multimodal potential of LLMs \cite{wang2024emu3,wu2024janus} enables seamless integration of textual and visual modalities, paving the way for unified generation frameworks. 

To demonstrate the effectiveness of our proposed hierarchical tokenizer, we choose the challenging task that autoregressive predict next token based LLM as our generative model. Our contributions are as follows:

\noindent \textbf{Design Hierarchical Video Tokenizer}: We design a multi-layer codebook that balances reconstruction quality and compression efficiency. By encoding videos at different granularities, high-level tokens capture semantic content, while lower-level tokens reconstruct spatiotemporal details.

\noindent \textbf{Improved Generation Performance and Compression Ratio}: We demonstrate that multi-layer tokenizers significantly improve generation performance compared to single-layer approaches. The high-level semantic tokens facilitate a smoother transition between textual and visual representations, leading to better alignment in text-to-video tasks.

\noindent \textbf{Dynamic Encoding and Masked Decoding}: Our framework introduces dynamic encoding mechanisms that reduce redundancy and enhance computational efficiency, leveraging masked decoding strategies to improve adaptability.

Our experiments validate the effectiveness of this approach, demonstrating that the hierarchical design not only achieves superior performance in both reconstruction and generation with high compression but also simplifies modeling for text-to-video generation by providing structured and semantically meaningful token representations.
\section{Related Work}
\label{sec:related_work}

\begin{table}
\caption{Video tokenizer compresss ratio comparison.}\label{tab: compress ratio}
\vspace{-3mm}
\begin{centering}
\scalebox{0.75}{
\begin{tabular}{cccc}
\hline 
Method & Input size & Num of tokens & Compress Ratio\tabularnewline
\hline 
MAGVIT-v2 \cite{yu2023magvitv2} & $17\times128\times128$ & 1280 & 217.6\tabularnewline
CogVideoX \cite{yang2024cogvideox} & - & - & 256\tabularnewline
EMU3 \cite{wang2024emu3} & $4\times512\times512$ & 4096 & 256\tabularnewline
\hline 
\textbf{Ours} & $64\times256\times256$ & 2448 & \textbf{1713.4}\tabularnewline
\hline 
\end{tabular}
}
\par\end{centering}
\end{table}

\subsection{Vision Tokenizer}

VQ-VAE \cite{van2017vqvae} and VQ-VAE-2 \cite{razavi2019vqvae-2}: The original VQ-VAE model introduced the concept of vector quantization for images, encoding continuous visual data into discrete codes that could be used in autoregressive modeling. VQ-VAE-2 extended this to hierarchical representations, improving the quality and resolution of generated images. This framework laid the groundwork for transforming images into discrete tokens that could be processed similarly to text.

DALL-E’s Discrete VAE (dVAE) \cite{ramesh2021dalle}: DALL-E utilized dVAE to tokenize images into discrete codes, leveraging the transformer architecture for image synthesis. This approach effectively represents complex visual concepts as tokens that can be manipulated in a structured manner, crucial for enabling vision-language models.

Video Tokenization Based on VQ-VAE: Building on VQ-VAE, VideoGPT \cite{yan2021videogpt} applied tokenization to video data, encoding each frame into discrete tokens to support autoregressive generation across temporal sequences. MAGVIT \cite{yu2023magvit} further refined this approach by introducing a masked generative video transformer, which encodes videos into multi-layer tokens, enhancing the representation of complex video scenes. MAGVIT-v2 \cite{yu2023magvitv2} improved upon MAGVIT by introducing a lookup-free quantization approach, enhancing reconstruction fidelity even in large vocabularies.

\subsection{Generation Based on Diffusion Models}
Diffusion models \cite{ho2020denoising,nichol2021improved,rombach2022stablediffusion} have become prominent in generative tasks for their ability to produce high-quality images \cite{saharia2022imagen,corneanu2024latentpaint} and videos \cite{blattmann2023stable,chen2024videocrafter2} through iterative stochastic denoising. Advanced architectures, such as DiT (Diffusion Transformer) \cite{peebles2023dit} and U-ViT (Unified Vision Transformer) \cite{bao2023u-dit}, have evolved from U-Net \cite{ronneberger2015unet}, offering scalability and enhanced modeling capacity for spatiotemporal data.

In text-to-video generation, Make-a-Video \cite{singer2022makeavideo} leverages pre-trained text-to-image diffusion models with video-only datasets, enabling video synthesis without paired text-video data. VDM \cite{ho2022vdm} incorporates spatiotemporal conditioning to ensure smooth temporal dynamics. Latent-Shift \cite{blattmann2023stable} extends Stable Diffusion into the temporal domain using latent space transformations for frame generation. VideoCrafter \cite{chen2024videocrafter2} builds on Stable Diffusion with a two-stage training strategy, introducing temporal consistency modules to enhance frame coherence. MicroCinema \cite{wang2024microcinema} explores efficient video generation with hierarchical latent spaces and coarse-to-fine refinement. Recent works, such as SORA\footnote{https://openai.com/index/video-generation-models-as-world-simulators/} \cite{brooks2024sora}, Vidu \cite{bao2024vidu}, and Lumiere \cite{bar2024lumiere}, demonstrate the potential of text to HD-long-video encoding. CogVideoX \cite{yang2024cogvideox} adopts a hierarchical coarse-to-fine strategy, integrating text-conditioned video token representations to improve semantic alignment.

% These methods highlight the versatility of diffusion-based models in capturing complex spatio-temporal dynamics and generating realistic, semantically aligned videos, paving the way for further advancements in text-to-video generation.

% image generation, video generation, from image, from text

\subsection{Generation Based on LLM Models}
Large Language Models (LLMs) \cite{touvron2023llama,achiam2023gpt} have recently emerged as an efficient alternative to diffusion models for generative tasks. Compared to diffusion models, LLMs offer faster sampling \cite{brown2020language}, better scalability to large datasets, and the ability to unify multimodal input-output tasks into a single framework \cite{yan2021videogpt,wang2024emu3}. While diffusion models rely on iterative denoising, LLMs leverage transformer architectures to predict tokens either autoregressively \cite{chen2020igpt} or via masked prediction \cite{devlin2018bert}, enabling structured and hierarchical generation.

Masked prediction models train LLMs to reconstruct missing tokens, enabling parallel training and efficient sampling. In image generation, MaskGit \cite{chang2022maskgit} introduced a masked token modeling strategy that iteratively refines images by progressively predicting missing tokens, achieving high-quality and computationally efficient synthesis. Extending this approach to video, MagViT-v2 \cite{yu2023magvitv2} builds on the MLM transformers \cite{devlin2018bert} used in MAGVIT \cite{yu2023magvit}, proposing an embedding method for iteratively masked video token modeling to support multi-task learning and enhance spatiotemporal consistency.

Autoregressive models predict tokens sequentially, modeling the conditional probability of each token given the previous ones. This paradigm, while more challenging due to its dependence on ordered token sequences, excels in capturing long-range semantic structure and producing highly coherent results. In the image domain, LlamaGen \cite{sun2024llamagen} utilized autoregressive transformers to generate semantically aligned and detailed images. Similarly, VAR \cite{tian2024var} employed a coarse-to-fine generation approach, progressively refining images across multiple levels of granularity. For video generation, recent works like VideoPoet \cite{kondratyuk2023videopoet} leverages MagViT-v2’s \cite{yu2023magvitv2} tokenizer to process multimodal inputs, including text, images, videos, and audio, demonstrating the versatility of autoregressive modeling in handling diverse modalities while maintaining spatiotemporal consistency. Show-O \cite{xie2024show} and EMU3 \cite{wang2024emu3} not only extend autoregressive models to text-to-video tasks but also introduce flexible frameworks that unify multimodal understanding and generation. These models seamlessly integrate text, image, and video inputs, enabling robust and coherent video synthesis aligned with textual descriptions.
\section{Method}
\label{sec:method}

\subsection{Rethinking about multi-modal generation}
% 重新思考多模态模型
% 经典的思考多模态视频生成模型训练pipeline通常是基于以下两个阶段，近期***方向展示了unified框架的潜力可以同时接受不同来自不同模态的token。然而此任务包含多个难点，视频数据具有 时间维度 和 空间维度 的复杂性，相较于图像数据，其连续帧之间的冗余和变化性大幅增加；视频编码需要在压缩冗余信息的同时保留必要的时空细节，而现有方法要么牺牲细节，要么导致过高的计算开销；下游text-to-video任务中，text token和visual token直接gap过大。所有的难点告诉我们需要去设计更好的video tokenizer，保证重建质量的情况下压缩比更大节省开销，同时下游任务中能更好的搭配language feature。
% 由于LLM相比于diffusion模型又以其灵活的输入输出展示了更高的灵活性，为了验证我们的hierarchy tokenizers在capture semantic informantion/bridge text-visual tokenizers我们选择了最难的长video的next token prediction autoregressive LLM模型作为下游验证，使用我们design的video tokenizer不仅仅在极低的bits per pixel（bpp）压缩比下获得了更好的reconstruction效果，相对于经典的single-layer video codebook，除了重建效果的提升，更明显的区别是当single codebook在完全生成失败的情况下，multi-layer tokenizers可以生成competitive results，能够很好的follow prompt里的关键词***。请注意我们的工作并不是为了contribute一个最好的text-to-video autoregressive LLM但是为了展示我们提出的高压缩比的tokenizer可以很好的作为模态间的过渡token，以及此结构的巨大潜力。后续的研究拥抱各种可能性，例如连续性而不是discrete tokens，结合diffusion model等。

Recent advances in multi-modal generative models show promise for unified framework \cite{wu2024janus,wang2024emu3}, and the typical pipeline framework consists of two main training stages:

% \begin{itemize} 
% \item \textbf{Stage 1}: Train visual autoencoder; 
% \item \textbf{Stage 2}: Train a generative model (e.g., diffusion or LLM) to predict understanding and generation results. 
% \end{itemize}

\noindent \textbf{Stage 1:} Train a visual autoencoder; 

\noindent \textbf{Stage 2:} Train a generative model (e.g., diffusion or LLM) to predict understanding and generation results. 

However, visual generation especially for video faces key challenges:  

\noindent \textbf{Complexity of Video Data:} Videos are inherently more complex than images, with spatial and temporal dimensions adding layers of redundancy and variability between consecutive frames. 

\noindent \textbf{Encoding Trade-Offs:} The autoencoder must balance detail preservation and computational efficiency, with designs tailored to specific use cases, though challenges may arise in certain downstream applications.

\noindent \textbf{Modality Gap:} Bridging the structural and semantic gap between text and video tokens remains a significant hurdle.  

These challenges highlight the need for better video autoencoder that balance high compression, reconstruction quality, and seamless text-video alignment for improved generative tasks.  

Compared to diffusion models, large language models (LLMs) exhibit flexibility in handling various input-output configurations, as shown in recent works \cite{yan2021videogpt,wu2024janus,wang2024emu3}. To validate the effectiveness of our hierarchical video tokenizer in bridging the gap between text and visual modalities, we adopt one of the most challenging downstream tasks: autoregressive next-token prediction for long video sequences using an LLM. 

Our experiments demonstrate that the hierarchical video tokenizer offers significant advantages in text-to-video generation, achieving high compression ratios while preserving superior reconstruction quality. Unlike single-layer codebooks, which often struggle under high compression, the multi-layer design produces coherent and competitive results, effectively capturing key semantic prompts. Additionally, hierarchical structure simplifies LLM modeling by providing compact visual tokens conditioned with text prompts, while residual dense tokens refine fine-grained details during generation. Rather than aiming to create the best autoregressive LLM with post-training modifications, this work emphasizes the potential of high-compression tokenizers as a robust intermediary for multi-modal tasks.  

Future research could explore continuity-based representations as an alternative to discrete tokens and integrate the hierarchical tokenizer with diffusion models or other generative approaches, paving the way for versatile and efficient multi-modal generative frameworks.

\subsection{Multi-layer discrete tokenizers design}

\begin{figure}[t]
\begin{center}
\includegraphics[width=1.0\linewidth]{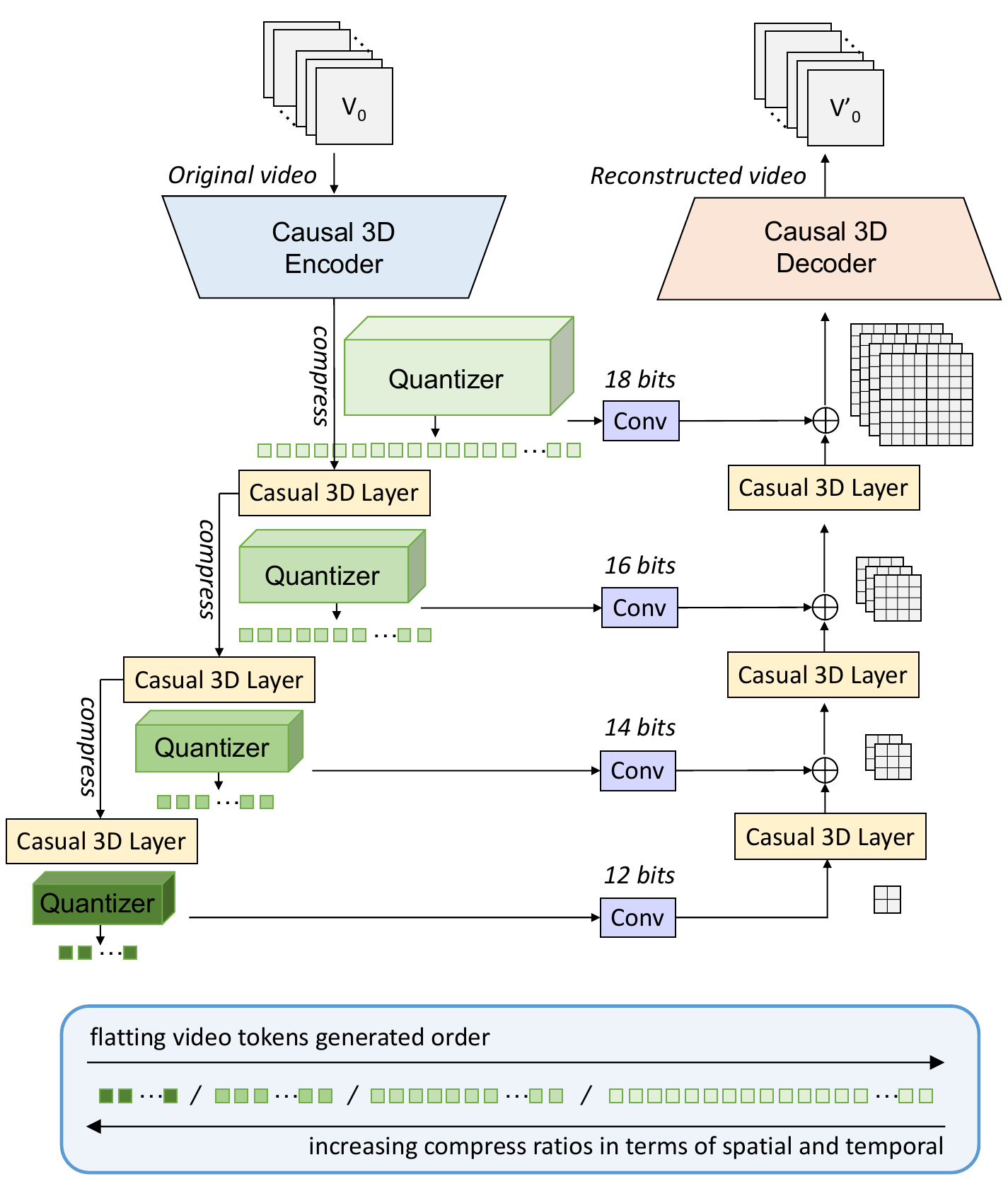}
\end{center}
\vspace{-5mm}
\caption{The overall architecture of \textbf{HiTVideo} tokenizer.}
\vspace{-3mm}
\label{fig: overall-vae}
\end{figure}

% Video data and compression ratio
\textbf{Tokenizer configuration:} Previous vision tokenizers based on VQ-VAE and VQ-GAN have shown the feasibility of generating images and videos from text prompts \cite{kondratyuk2023videopoet, tian2024var, sun2024llamagen} and demonstrated potential for unified cross-modality token understanding \cite{wang2024emu3}. Among recent advancements, MAGVIT-v2 \cite{yu2023magvitv2} stands out as a highly advanced video tokenizer, achieving high-quality reconstruction by utilizing a large vocabulary size with lookup-free quantization (LFQ). MAGVIT-v2 compresses only 17 frames (approximately 2.125 seconds) of $128 \times 128$ resolution video into a $5 \times 16 \times 16$ latent representation, flattened to 1280 tokens. This relatively low compression ratio limits the efficiency of redundant information handling, making the generation of long video sequences computationally expensive when using large language models (LLMs) as the generative framework. Additionally, encoding only 17 frames at a time restricts the tokenization model’s capacity to capture the continuity essential for comprehensive video scenes. For instance, methods like VideoPoet \cite{yan2021videogpt} also tokenize brief 17-frame sequences, which results in challenges when attempting to maintain narrative consistency across multiple segments, thus complicating the task of producing cohesive and complete video narratives from short, disjointed shots.

To address these limitations, we draw inspiration from the average scene shot duration in film production and propose a video tokenizer capable of handling extended video sequences. Specifically, our approach encodes 64 frames at 8 frames per second, spanning 8 seconds, to provide a broader temporal context for capturing scene-level coherence and continuity. This longer-duration tokenizer is designed to process input sequences that better represent complete scenes, effectively capturing temporal dependencies and narrative flow. Our design prioritizes maintaining high reconstruction quality while achieving a higher compression ratio, enabling more efficient long-sequence video generation. The detailed compression configurations for the video tokenizer are provided in Tab.~\ref{tab: compress ratio}.

\begin{figure*}[t]
\begin{center}
\includegraphics[width=1.0\linewidth]{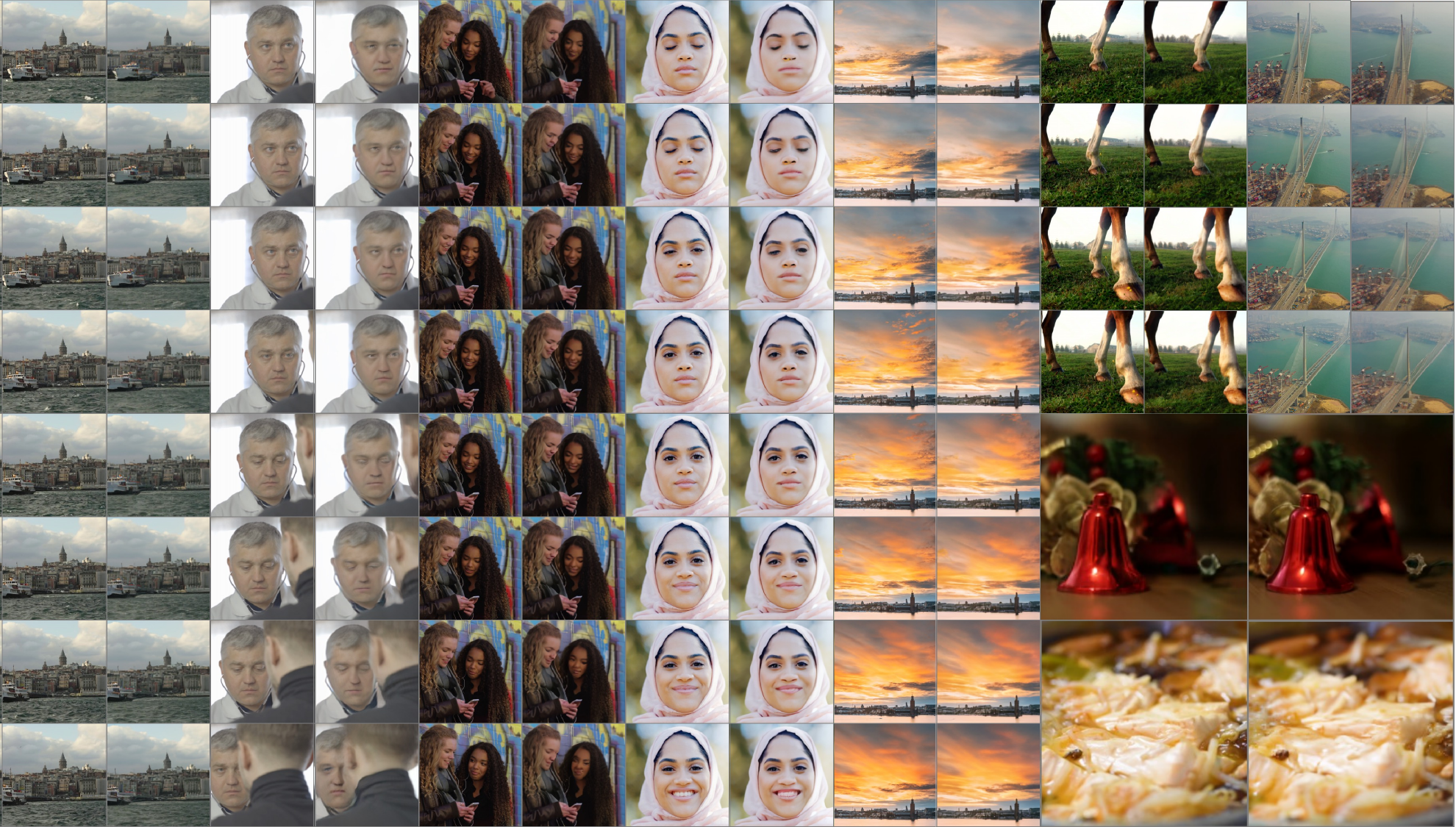}
\end{center}
\vspace{-5mm}
\caption{Qualitative evaluation of our proposed hierarchical tokenizers. In each case, the left column represent partial frames sampled from the 64 input video, while the right column presents the corresponding reconstruction results.}
\vspace{-3mm}
\label{fig: stage1_vis}
\end{figure*}

% Question: do you need to mention about the single codebook design?
\noindent \textbf{Architecture:} Our approach builds on the foundational 3D VAE architecture used in VideoPoet \cite{yan2021videogpt}. This model employs a causal 3D encoder for down-sampling across spatial and temporal dimensions, paired with a symmetrical causal 3D decoder for up-sampling. These components, together with a quantizer (e.g., a look-up mechanism for regularization), enable efficient encoding. Inspired by advancements in 2D VAEs for image encoding \cite{esser2021taming,rombach2022stablediffusion}, 3D VAEs achieve considerably higher compression ratios \cite{yang2024cogvideox}, which improves both video quality and temporal consistency in reconstructions. The use of causal 3D CNNs further strengthens our model’s ability to generate coherent, visually accurate video sequences \cite{yu2023magvitv2}. The detailed architecture can be found in Fig.~\ref{fig: overall-vae}. 

The primary 3D causal encoder and decoder in our model consist of four ResNet blocks, each equipped with causal 3D CNN layers along the temporal dimension. For both compression in the encoder and uncompress in the decoder, spatial processing is applied at each of the four block layers, while temporal processing occurs specifically within the first three block layers. This hierarchical down-sampling approach enables efficient compression across both spatial and temporal dimensions. More specifically, starting from an input video resolution of $64\times256\times256$, the encoder progressively reduces the video data to a latent size of $8\times16\times16$. Compared to other methods \cite{yu2023magvitv2, yang2024cogvideox, wang2024emu3}, our approach achieves an additional 8x compression when combining both spatial and temporal dimensions. This allows for a substantial reduction in data storage requirements while still retaining essential spatiotemporal information. 

Although our tokenizer achieves high-quality reconstruction of long videos at a high compression ratio, it encounters sampling failures in generation tasks based on large language models (LLMs). We attribute this to a substantial semantic gap between vision tokens, which are optimized for visual reconstruction, and text tokens, which capture abstract semantic information. Inspired by VAR \cite{tian2024var}, which demonstrated the effectiveness of a coarse-to-fine approach in image generation using a GPT-style autoregressive model conditioned on class embeddings or text, we adopted a hierarchical strategy similar to VQ-VAE-2 \cite{razavi2019vqvae-2}. Following the main 3D causal encoder, our approach progressively compresses latent features with lightweight causal 3D CNN layers. Notably, as the latent space is further compressed, the quantization dim in LFQ is reduced accordingly. In the decoder phase, a symmetrical network structure is introduced to progressively reconstruct the video content.

Our hierarchical design integrates multiple codebooks, each operating at a distinct level of granularity. The highest layers compress across all 64 frames, capturing broader semantic information, while the lower layers focus on precise visual details to enhance reconstruction quality. This structure aims to bridge the semantic gap between text and vision tokens by initially generating a rough layout aligned with textual semantics, then incrementally refining it to achieve coherent and high-fidelity video outputs.

\noindent \textbf{Loss Function:} Our tokenizer is trained jointly using a combination of L1 loss, LPIPS perceptual loss \cite{zhang2018LPIPS}, and adversarial loss \cite{goodfellow2014gan} with a discriminator to enhance reconstruction quality. Following MAGVIT-v2 \cite{yu2023magvitv2}, we also incorporate an entropy penalty to encourage effective vocabulary utilization across each codebook layer. To facilitate more effective training of the multi-layer codebook, we adopt a progressive training strategy. This approach initially focuses on training with higher-compression codebooks to reconstruct a coarse version of the video, gradually adjusting and refining the model on denser codebooks, thereby enhancing the detail in reconstructed outputs.

% By leveraging this multi-layer tokenizer framework, our model can more effectively balance reconstruction quality with semantic coherence, ensuring that the generated video content accurately reflects the provided text prompts while maintaining high visual fidelity.

\begin{figure*}[t]
\begin{center}
\includegraphics[width=1.0\linewidth]{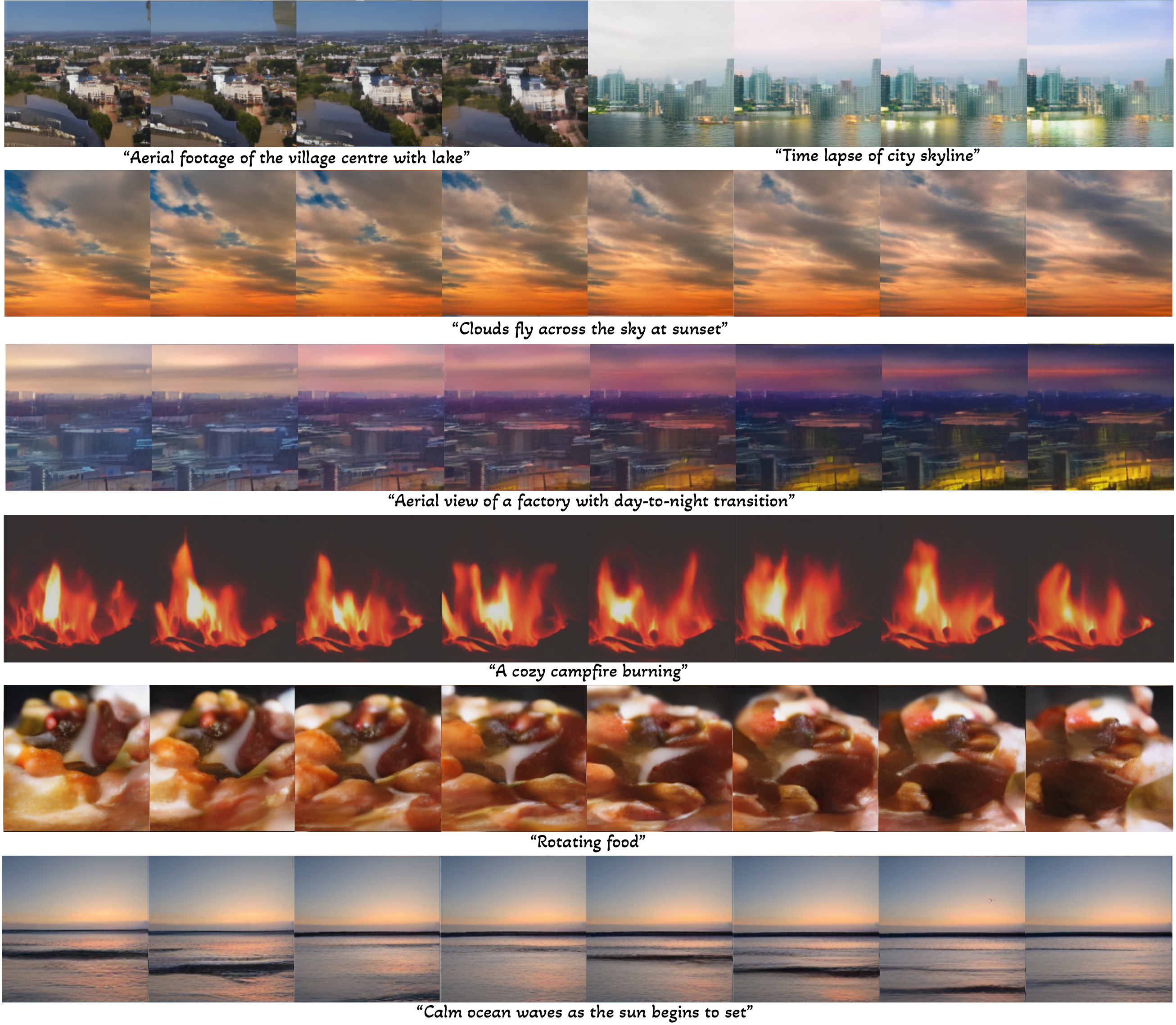}
\end{center}
\vspace{-8mm}
\caption{Qualitative generative results of our HiTVideo tokenizers.}
\vspace{-5mm}
\label{fig: stage2_vis}
\end{figure*}

\subsection{Text-to-video Generation by Autoregressive LLM}

We utilize the Llama-3B model \cite{touvron2023llama,sun2024llamagen} as the large language model (LLM) to predict the next token, training it to interpret and generate content based on the text prompt. The text prompt is first processed by a frozen Flan-T5-XL model \cite{chung2024flan} to extract text embeddings, which serve as conditioning input for the autoregressive model. The video tokens are then appended after the initial text tokens with right padding applied. Inspired by the animation creation process, where animators start with high-level storyboards and progressively refine details, the sequence of video tokens begins with the most compressed (high-level) tokens and gradually transitions to denser tokens. This ordered sequence of tokens are shown in the bottom of Fig.~\ref{fig: overall-vae} and can be represented as:
\vspace{-1mm}

\[
p(v | t) = \prod_{m=1}^{M} \prod_{n=1}^{N^m} p(v_n^m \mid t, v_{<n}^{\leq m})
\]
where $m$ represents the $m-$th tokenizer layer and $N^m$ denotes the total number of tokens for the $m-$th vocabulary, the next token $v_{n}^{m}$ is generated based on previous tokens $v_{<n}^{\leq m}$, as well as all text condition tokens $t$.

Rotary Position Embedding (RoPE) \cite{su2024rope} effectively captures relative positional information between tokens and excels in modeling long token sequences \cite{yang2024cogvideox}, outperforming learnable absolute position embeddings. To represent the positions of video tokens across multi-layer vocabularies in spatial and temporal dimensions, we extend 2D RoPE embeddings to 3D, allocating feature dimensions across three axes. Each layer’s codebook applies position encoding independently, while layer-specific positional encodings are incorporated to inform the GPT model about the layer, temporal frame, and spatial location of the token being predicted. Inspired by diffusion models, we include learnable unconditional embeddings to support classifier-free guidance (CFG) generation \cite{ho2022classifier,dhariwal2021diffusion} to improve diversity. 
% Details of the architecture are provided in Tab.~\ref{tab: archi_ar}.

During training, we use shifted supervision to predict the next token, applying cross-entropy loss independently for the classification logits of multi-layer vocabularies. To ensure balanced learning across hierarchical layers, distinct loss weights are assigned to each layer. During inference, condition text embeddings are used to prefill the first token, and key-value caching (KV-cache) mechanism \cite{dai2019transformer,pope2023efficiently} is employed with positional input to optimize memory usage and reduce computational overhead. The CFG factor is incorporated to balance diversity and fidelity during generation. This framework allows us to sample video sequences of up to 64 frames efficiently, maintaining an acceptable speed for long sequences. It is important to note that our primary objective is to demonstrate the benefits of hierarchical tokenizers for simplifying the modeling complexity of video generation with LLM. We do not apply post-training techniques, such as fine-tuning on aesthetic datasets or super-resolution, to further enhance inference results.

\begin{table}
\caption{The detailed configuration of our autoregressive model.}
\vspace{-3mm}
\label{tab: archi_ar}
\centering{}%
\scalebox{0.8}{
\begin{tabular}{cc}
\hline 
Setting & Detail\tabularnewline
\hline 
Parameters & 3.1B\tabularnewline
Layers & 24\tabularnewline
Hidden Size & 3200\tabularnewline
Heads & 32\tabularnewline
Text Length & 120\tabularnewline
Residual Dropout & 0.1\tabularnewline
FFN Dropout & 0.1\tabularnewline
Token Dropout & 0.1\tabularnewline
\hline 
\end{tabular}
}
\end{table}

\section{Experiments}
\label{sec:experiments}

\subsection{Video Tokenizers}

\noindent \textbf{Implementation details:} We train our hierarchical tokenizers as architecture details in Tab.~\ref{tab: archi_ar} on the Pexels dataset\footnote{\url{https://www.pexels.com/search/videos/videos}} at a resolution of $256\times256$ and 8 FPS. To assess the effectiveness of our method, we evaluate the bits per pixel (bpp) alongside reconstruction metrics, including LPIPS \cite{zhang2018LPIPS}, PSNR \cite{jahne2005psnr}, SSIM \cite{wang2004ssim}, and MS-SSIM \cite{wang2003multiscale}, on the validation set, as summarized in Tab.\ref{tab: stage1_metric}. Qualitative comparisons are provided in Fig.\ref{fig: stage1_vis}. Compared to prior methods, our proposed video tokenizer achieves state-of-the-art reconstruction quality with a 70\% reduction in bits per pixel. Tab.~\ref{tab: stage1_metric} demonstrates its ability to balance compression and reconstruction for long video sequences. Subsequent experiments further show improved multimodal alignment during generation compared to single-layer dense tokenizers at low compression ratios. Furthermore, ablation study on the tokenizer design reveals several key insights and potentials:

\begin{table}
\vspace{-3mm}
\caption{Video tokenizers comparison.}\label{tab: stage1_metric}
\vspace{-3mm}
\begin{centering}
\scalebox{0.76}{
\begin{tabular}{cccccc}
\hline 
method & bpp & LPIPS$\downarrow$ & PSNR$\uparrow$ & SSIM$\uparrow$ & MS-SSIM$\uparrow$ \tabularnewline
\hline 
\rowcolor{gray!20} HEVC(H.265)\cite{sullivan2012hevc} & 0.3$-$0.6 & 0.199 & 30.10 & - & 0.943 \tabularnewline
\rowcolor{gray!20} VVC (H.266)\cite{bross2021vvc} & 0.2$-$0.4 & 0.153 & 32.65 & - & 0.966 \tabularnewline
\hline 
MAGVIT\cite{yu2023magvit} & 0.0384 & 0.144 & 23.70 & - & 0.846\tabularnewline
MAGVIT-v2\cite{yu2023magvitv2} & 0.0384 & \textbf{0.104} & 26.18 & - & 0.894\tabularnewline
EMU3\cite{wang2024emu3} & 0.0384 & 0.109 & 21.59 & 0.622 & - \tabularnewline
\hline 
\textbf{Ours} & \textbf{0.0120} & 0.108 & \textbf{27.53} & \textbf{0.791} & \textbf{0.907}\tabularnewline
\hline 
\end{tabular}
}
\par\end{centering}
\end{table}

\begin{table*}
\caption{Configuration details for HiTVideo tokenizer at 256 resolution and 64 frames per video. Left table: multi-layer codebook structure. Right table: single-layer structure, with latent size interpolated to ensure a similar token count per video.}\label{tab: archi_stage1}
\vspace{-3mm}

\centering{}%

\scalebox{0.8}{
\begin{tabular}{ccccc}
\hline 
\multirow{2}{*}{Config} & \multicolumn{4}{c}{Layer Index}\tabularnewline
\cline{2-5} \cline{3-5} \cline{4-5} \cline{5-5} 
 & 0 & 1 & 2 & 3\tabularnewline
\hline 
Codebook Size & $2^{18}=262144$ & $2^{16}=65536$ & $2^{14}=16384$ & $2^{12}=4096$\tabularnewline
Compression  & $8\times8\times8$ & $16\times32\times32$ & $32\times32\times32$ & $64\times64\times64$\tabularnewline
Latent Size & $8\times16\times16$ & $4\times8\times8$ & $2\times8\times8$ & $1\times4\times4$\tabularnewline
Quant Dim & 18 & 16 & 14 & 3\tabularnewline
\hline 
Number of Tokens & 2048 & 256 & 128 & 16\tabularnewline
\hline 
\end{tabular} %
\begin{tabular}{cc}
\hline 
\multirow{2}{*}{Config} & Layer Index\tabularnewline
\cline{2-2} 
 & 0\tabularnewline
\hline 
Codebook Size & $2^{18}=262144$\tabularnewline
Compression  & $8\times8\times8$\tabularnewline
Latent Size & $8\times17\times17$\tabularnewline
Quant Dim & 18\tabularnewline
\hline 
Number of Tokens & 2312\tabularnewline
\hline 
\end{tabular}
}
\end{table*}

\begin{figure}[t]
\begin{center}
\includegraphics[width=1.0\linewidth]{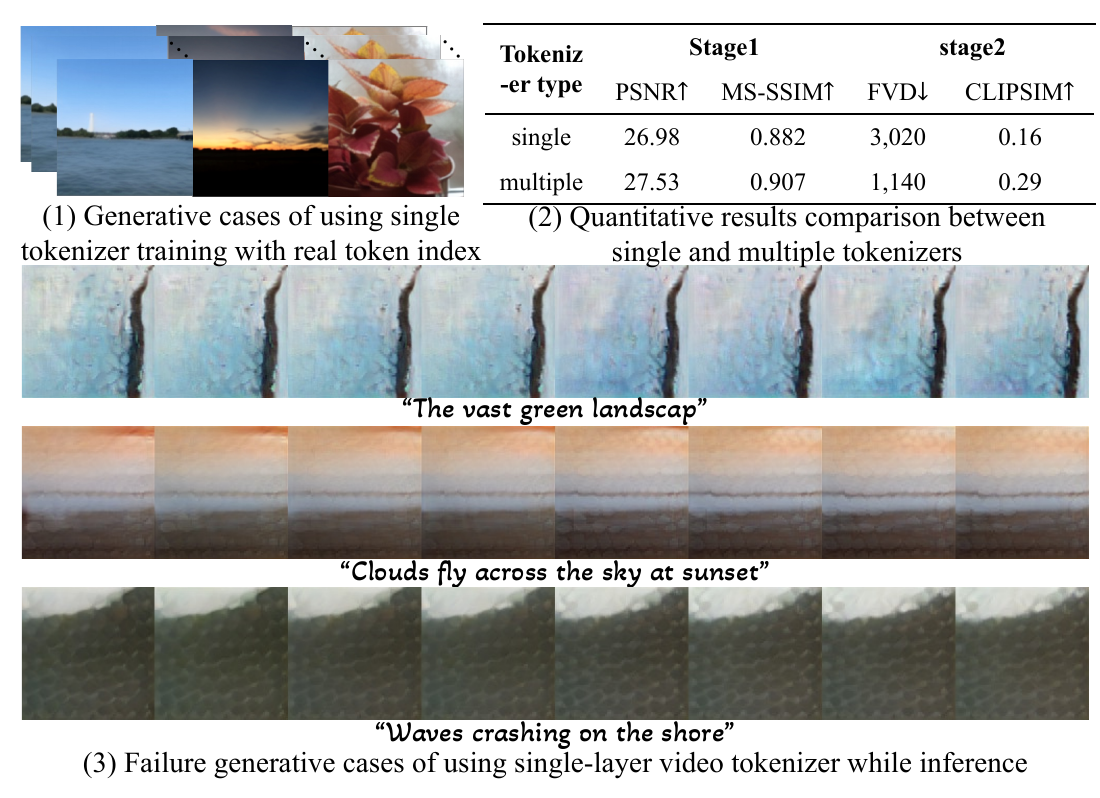}
\end{center}
\vspace{-5mm}
\caption{Comparison of using single and multi-layer video tokenizers as $64\times256\times256$ video resolution.}
\vspace{-5mm}
\label{fig: singlevsmultiple}
\end{figure}

\noindent \textbf{Improvements with hierarchical tokenizers:} For a fair comparison, we maintain the total token count consistent with the single-layer setup by interpolating the spatial dimensions before downsampling. The detailed compression configurations are provided in Tab.~\ref{tab: archi_stage1}. Multi-layer codebooks achieve superior reconstruction quality compared to single-layer codebooks. The quantitative comparison of reconstruction quality on the Pexels dataset is presented inas shown in Fig.~\ref{fig: singlevsmultiple}-(2). Beyond reconstruction, multi-layer codebooks show significant advantages in autoregressive generation tasks, as highlighted in Fig.~\ref{fig: singlevsmultiple}. Notably, using a dense single-layer codebook, the LLM fails to generate a coherent 64-frame video from a language prompt as shown in Fig.~\ref{fig: singlevsmultiple}-(3). This underscores the potential of multi-layer structures not only for efficient compression but for bridging the semantic gap between language and video tokens.

\begin{figure}[h]
\begin{center}
\includegraphics[width=1.0\linewidth]{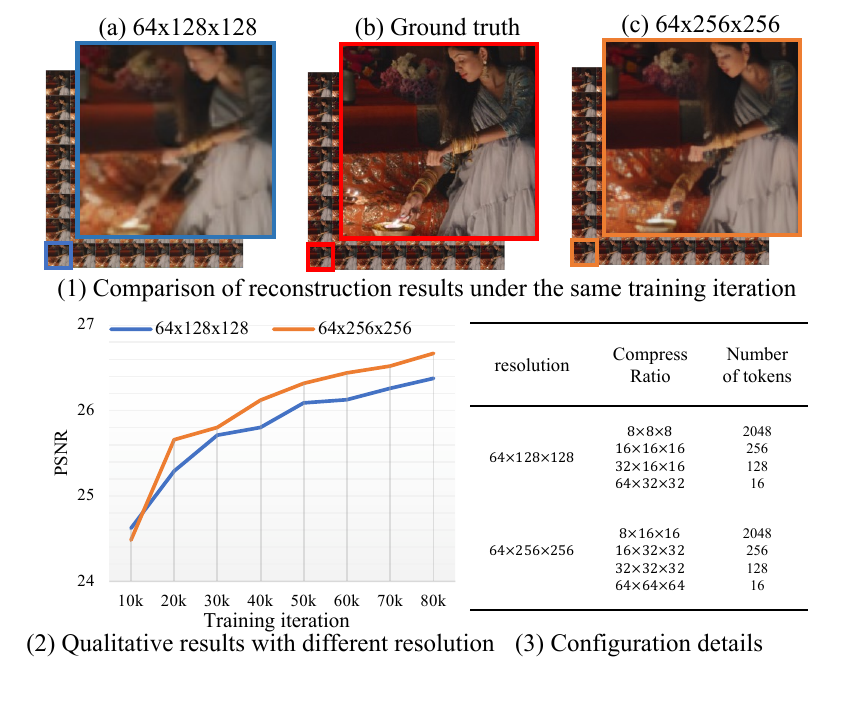}
\end{center}
\vspace{-5mm}
\caption{Comparison of qualitative and quantitative results across different input resolutions with same total number of video tokens.}
\vspace{-3mm}
\label{fig: 128vs256}
\end{figure}

\noindent \textbf{Higher Resolution with Higher Compression Ratios Improves Performance:} For a fixed total number of video tokens, we observe that utilizing higher input resolutions with larger compression ratios enables the model to capture finer details within the same number of training iterations. This approach ensures that, despite significant compression, the higher-resolution input preserves more spatial information, leading to enhanced reconstruction quality. To ensure a fair comparison, both the 128-resolution and 256-resolution models were initialized with identical parameters and quantization strategies. As illustrated in Fig.~\ref{fig: 128vs256}, the higher-resolution model demonstrates faster convergence in terms of PSNR, indicating superior reconstruction quality. At the same training iteration, videos reconstructed by the higher-resolution model exhibit greater visual fidelity and more detailed features, closely resembling the ground truth, underscoring the advantages of this strategy under high compression settings.

\begin{figure}[t]
\begin{center}
\includegraphics[width=1.0\linewidth]{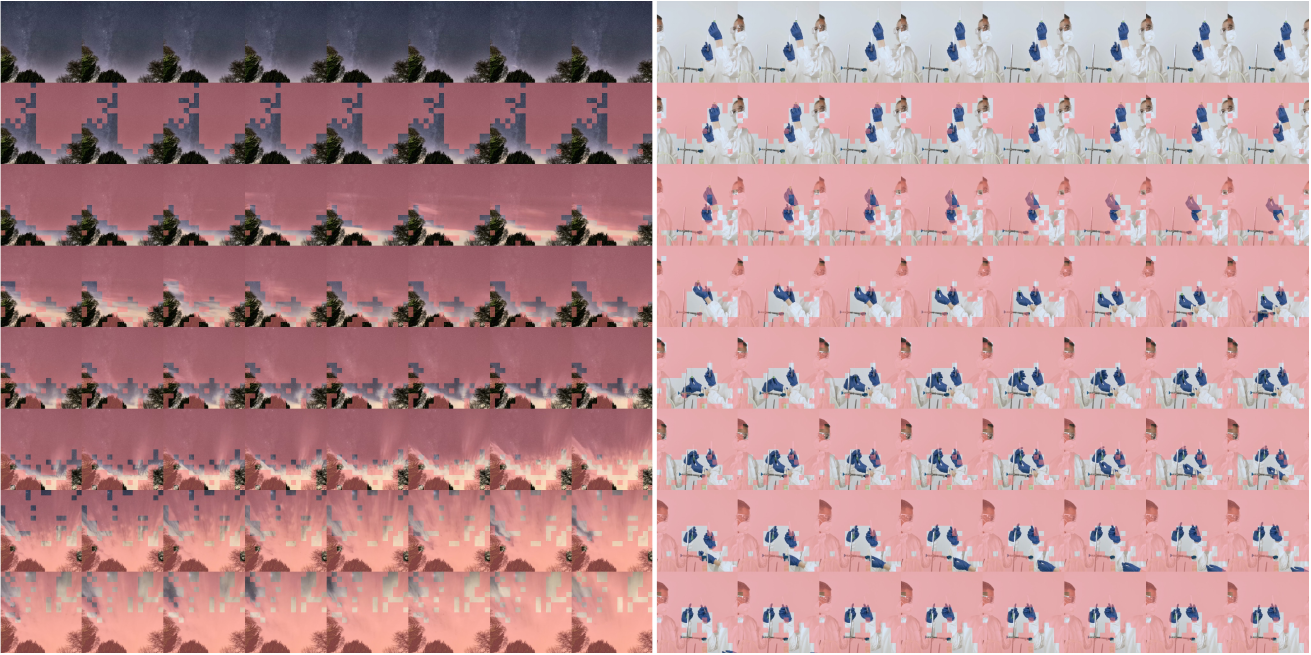}
\end{center}
\vspace{-5mm}
\caption{Visualization of redundant quantized latents.}
\vspace{-5mm}
\label{fig: mask_vis}
\end{figure}

\noindent \textbf{Dynamic video encoding with hierarchical tokenizers:} As shown in Tab.~\ref{tab: archi_stage1}, dense-layer tokenization generates 2048 tokens per 64-frame video, accounting for approximately 84\% of the total token budget and incurring high computational costs. Through VAE pretraining with multi-layer codebooks, we observed that consecutive frames often exhibit high similarity after quantization, indicating redundancy in dense tokenizers due to the inherent properties of video data (illustrated in Fig.~\ref{fig: mask_vis}). To quantify this redundancy, we compute the differential matrix across video frames. The pink mask highlights video tokens with difference scores below the matrix average, identifying redundant patches within the video.

\begin{figure}[t]
\begin{center}
\includegraphics[width=1.0\linewidth]{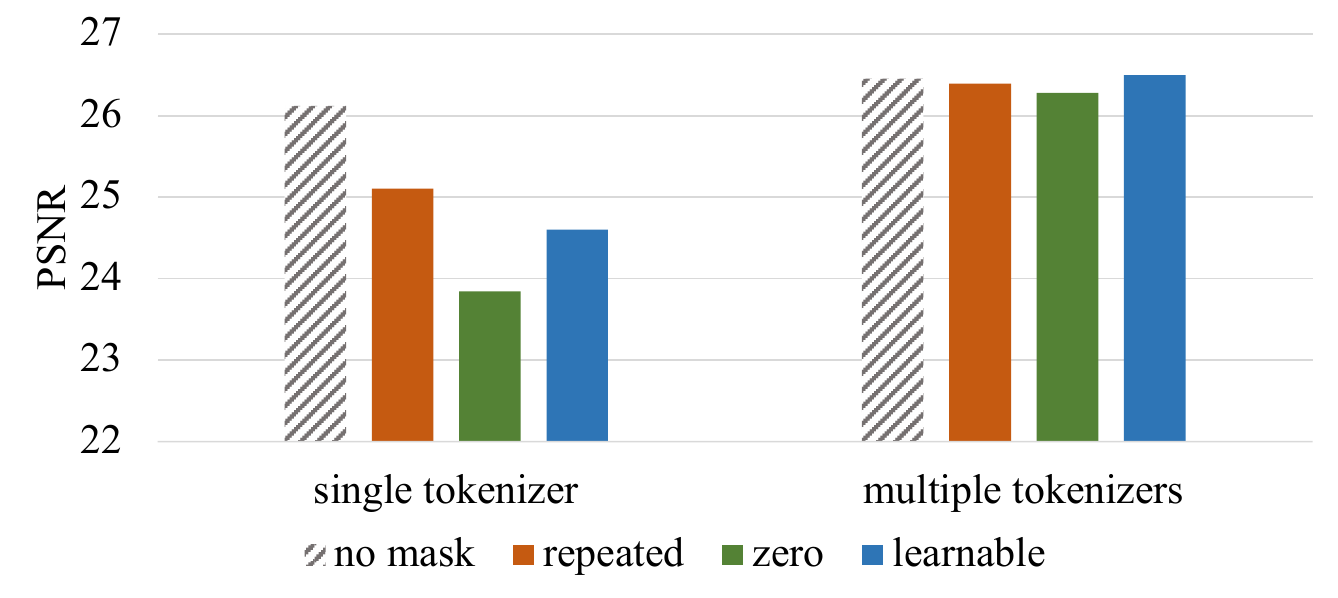}
\end{center}
\vspace{-5mm}
\caption{PSNR results while masked decoding.}
\vspace{-5mm}
\label{fig: maskdecoder}
\end{figure}

% Our hierarchical tokenizer design enables dynamic video encoding by masking these redundant patches. 
We apply three masking strategies: repeated tokens from the previous frame, zero padding tokens, and learned mask tokens. To balance static and dynamic content, we limit the masking ratio to 85\%, ensuring a minimum 15\% unmasked patches by random sampling if necessary. As shown in Fig.~\ref{fig: maskdecoder}, all masking strategies allow video reconstruction, with scores reported for the same training iterations. Notably, single codebooks with mask significantly degrade reconstruction performance, highlighting the potential of hierarchical tokenizers for dynamic compression and efficient video generation with fewer tokens.

\subsection{Text-to-video Generation}

% \noindent \textbf{Implementation Details:} 
To demonstrate the effectiveness of our hierarchical tokenizers in generative models, we trained the Llama-3B model to predict the next token autoregressively on Pexels dataset. During inference, the first token is prefilled according to text-conditioned embeddings, and unconditioned embeddings are adjusted using a classifier-free guidance (CFG) scalar. 
% The top-k sampling parameter is set to 1,000, and the temperature is set to 1.0. 
Qualitative results are shown in Fig.~\ref{fig: stage2_vis}, while quantitative performance on Pexels prompts is evaluated using FVD \cite{unterthiner2018fvd} and CLIPSIM \cite{radford2021clip}, as reported in Fig.~\ref{fig: singlevsmultiple}-(2).

 For a fair comparison, we trained the autoregressive LLM using a pre-trained single-layer tokenizer on 64 frames. While the model successfully sampled video tokens during training when provided with ground truth causal token indices, as shown in Fig.~\ref{fig: singlevsmultiple}-(1), it failed to generate coherent and meaningful videos when conditioned solely on text prompts, as illustrated in Fig.~\ref{fig: singlevsmultiple}-(2-3). These findings highlight the effectiveness of our highly compressed hierarchical tokenizers in bridging the semantic gap between text and dense visual tokens, thereby facilitating robust text-to-video generation.

\section{Conclusion}
\label{sec:conclusion}
Text-to-video generation remains a challenging yet exciting frontier in generation research. Inspired by traditional animation workflows and prior research, we propose a multi-layer codebook video tokenizer that effectively balances compression efficiency and reconstruction quality. By capturing high-level semantic content while preserving essential spatiotemporal details, our approach enhances the quality of generated video content under high compression. Compared to single-layer tokenization, our hierarchical design improves reconstruction fidelity while simplifying LLM modeling by providing a more abstract and semantically meaningful representation. Increasing the compression ratio by integrating dynamic encoding with a mask-based decoder further unlocks the potential of our design. Limitation: We leave the integration of our hierarchical tokenizers with masked LLMs or diffusion models and multi-modality framework with understanding task as directions for future exploration.

{
    \small
    \bibliographystyle{ieeenat_fullname}
    \bibliography{main}
}

% WARNING: do not forget to delete the supplementary pages from your submission 

\clearpage
\setcounter{page}{1}
% \maketitlesupplementary
\appendix
\twocolumn[
  \centering
  \section*{Appendix}
  \addcontentsline{toc}{section}{Appendix}
  \vspace{5mm}
]

\vspace{10mm}

\section{Effect of CFG Scale in Text-to-Video Generation}

The selection of an appropriate classifier-free guidance (CFG) scale is a crucial factor influencing the quality of generated videos. Originally proposed in the context of diffusion models, the CFG scale determines the trade-off between fidelity to the textual prompt and the diversity of generated visual content. Classifier-free guidance integrates conditional (prompt-based) and unconditional (prompt-free) signals during model training, thereby enhancing the model's capacity to adhere to the provided textual descriptions without sacrificing model's generative flexibility.

Specifically, the CFG scale directly modulates this balance. Higher CFG values enforce strict adherence to the input prompt, yielding videos that are semantically precise and visually coherent. However, an excessively high CFG scale can limit the output diversity, resulting in repetitive or overly rigid generations. Conversely, lower CFG scales offer greater creative freedom, allowing the model to explore diverse interpretations beyond the prompt, thus enhancing the variability and naturalness of generated videos, albeit potentially reducing semantic accuracy.

Inspired by the CFG mechanism used in diffusion models, we extend this concept to an autoregressive language-model-based text-to-video generation framework. During training, we introduce an unconditional embedding, enabling the model to effectively utilize conditional and unconditional contexts simultaneously. This strategy equips the model with greater flexibility, allowing it to generate outputs that remain semantically coherent while exploring creatively diverse variations. To empirically investigate the effect of the CFG scale, we conducted experiments across various CFG configurations, examining the inherent trade-offs between semantic alignment and visual diversity. 

The balance determined by the CFG scale is pivotal, directly affecting both the semantic coherence and the visual appeal of the generated videos. Therefore, careful tuning of the CFG parameter is essential for optimizing video generation quality. The computation of logits used for sampling video tokens, incorporating the CFG factor, can be represented by the following pseudo-code:

\begin{align*}
\mathtt{logits} &= \mathtt{uncond\_logits} + \\
&\quad (\mathtt{cond\_logits} - \mathtt{uncond\_logits}) \cdot \mathtt{cfg\_scale}
\end{align*}

\begin{figure*}[hb]
\begin{center}
\includegraphics[width=1.0\linewidth]{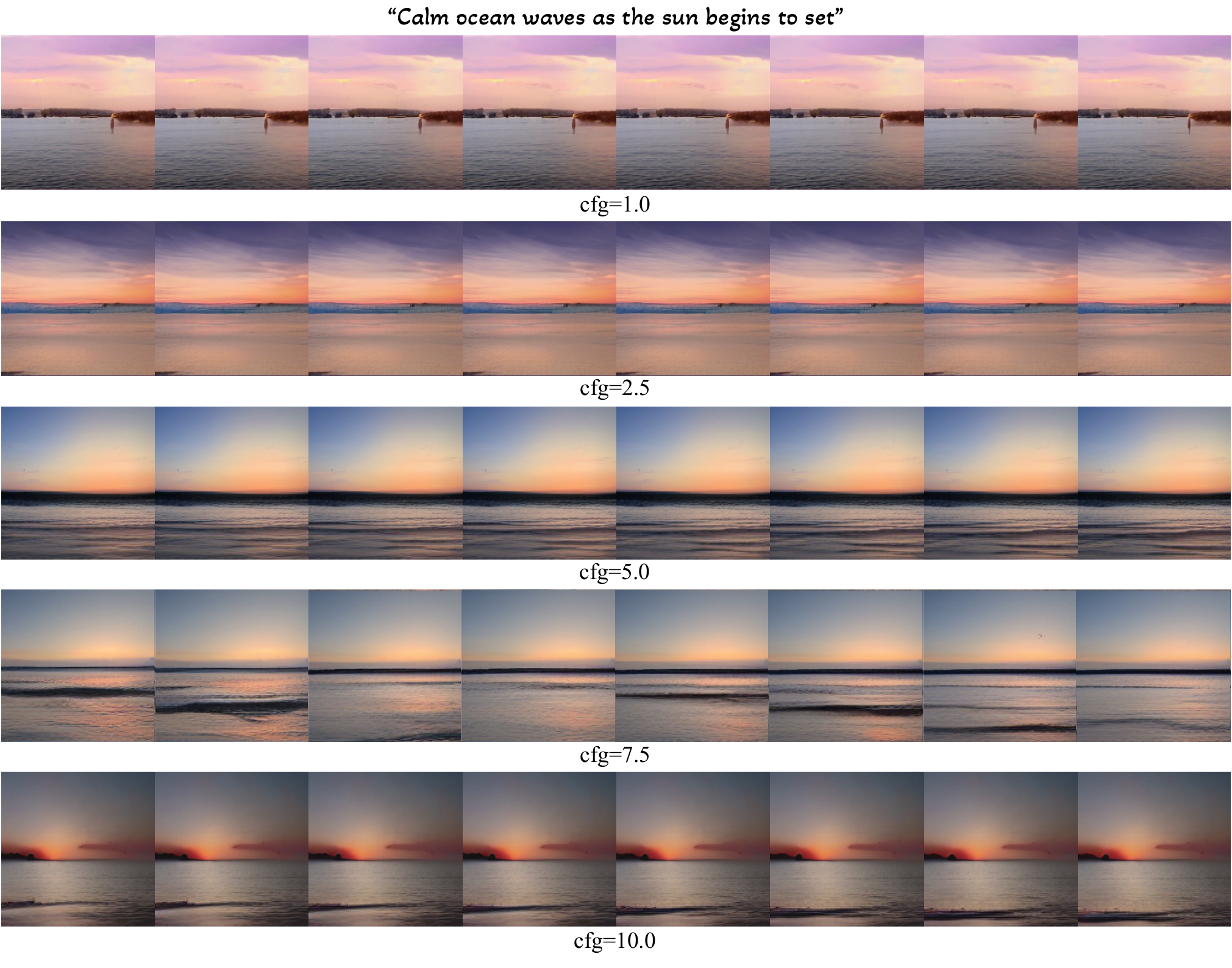}
\end{center}
\vspace{-5mm}
\caption{Generative results of case 1 with different cfg factor.}
\vspace{-5mm}
\label{fig: cfg_1}
\end{figure*}

\begin{figure*}[hb]
\begin{center}
\includegraphics[width=1.0\linewidth]{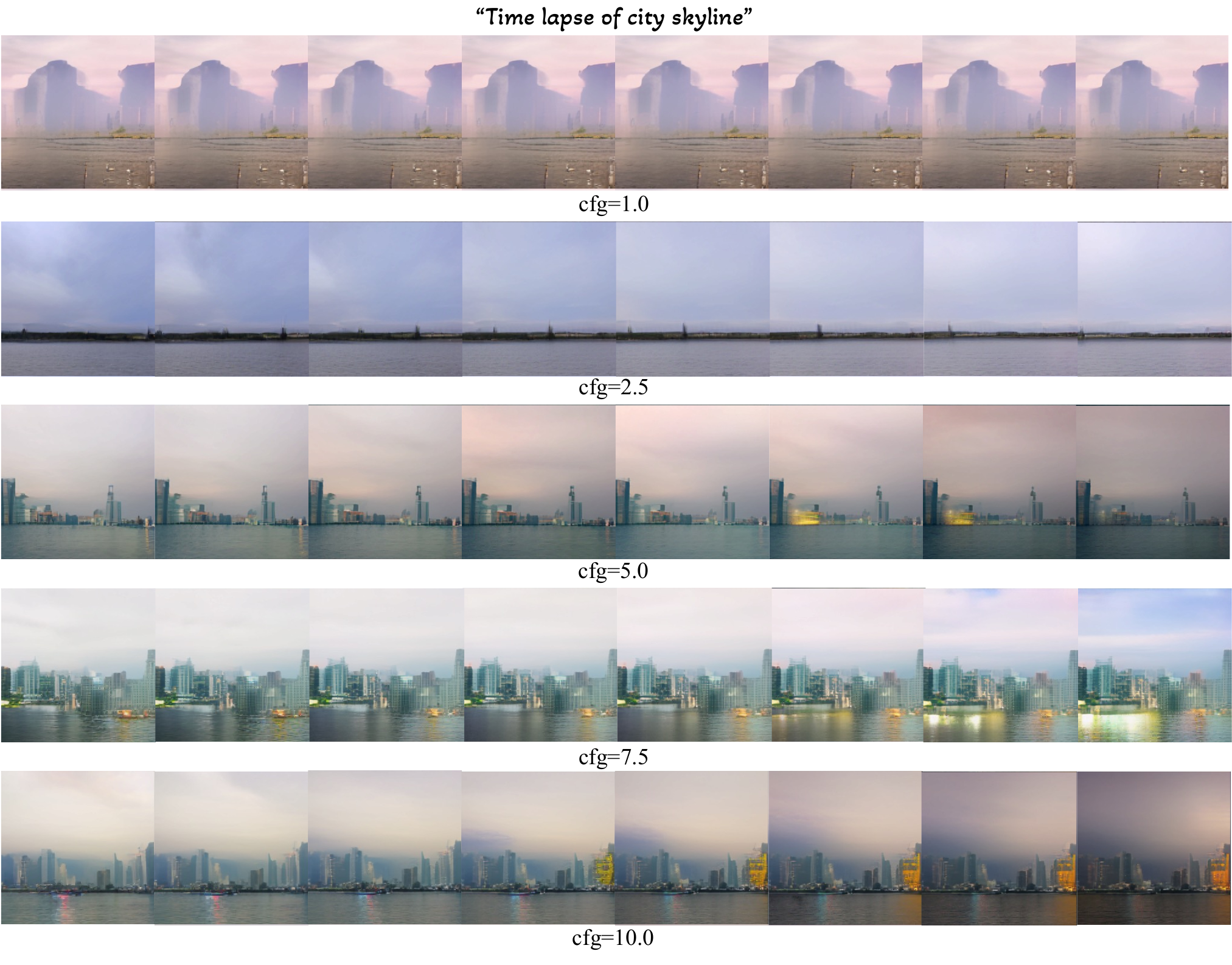}
\end{center}
\vspace{-5mm}
\caption{Generative results case 2 with different cfg factor.}
\vspace{-5mm}
\label{fig: cfg_2}
\end{figure*}

\begin{figure*}[t]
\begin{center}
\includegraphics[width=1.0\linewidth]{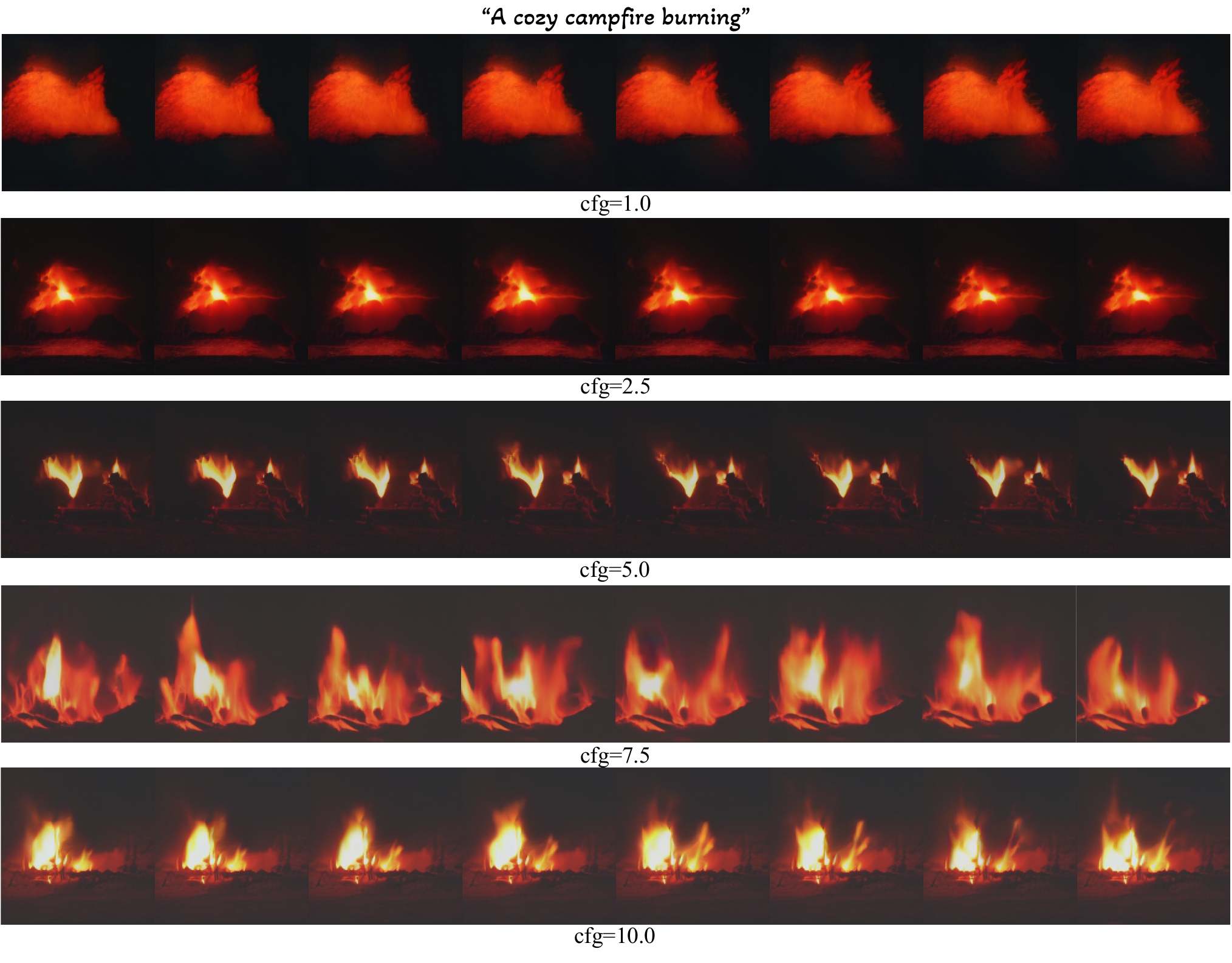}
\end{center}
\vspace{-5mm}
\caption{Generative results case 3 with different cfg factor.}
\vspace{-5mm}
\label{fig: cfg_3}
\end{figure*}

In our experiments, we observed that CFG scales within the range of 5.0 to 7.5 typically result in high-quality videos with optimal adherence to input prompts, as illustrated in Fig.~\ref{fig: cfg_1} - Fig.~\ref{fig: cfg_3}. Consequently, we employ a CFG scale of 7.5 as the default setting during inference.

\section{Analysis of Diversity in Text-to-Video Generation Relative to the Training Samples from Pexels Dataset}

To evaluate the diversity of generated videos, we compare model outputs against corresponding videos from the Pexels training dataset, conditioned on identical textual captions. This comparative analysis demonstrates the model's capability to generate novel and diverse content while maintaining semantic consistency with the input captions.

For each caption, multiple video samples are generated by varying the random seed. This strategy enables us to comprehensively examine the variability introduced by the model into its outputs. Such analysis reveals the model's ability to transcend mere replication of training samples, highlighting its potential for producing creatively diverse and visually distinctive outputs, essential qualities in generative video modeling.

\begin{figure*}[t]
\begin{center}
\vspace{-10mm}
\includegraphics[width=0.97\linewidth]{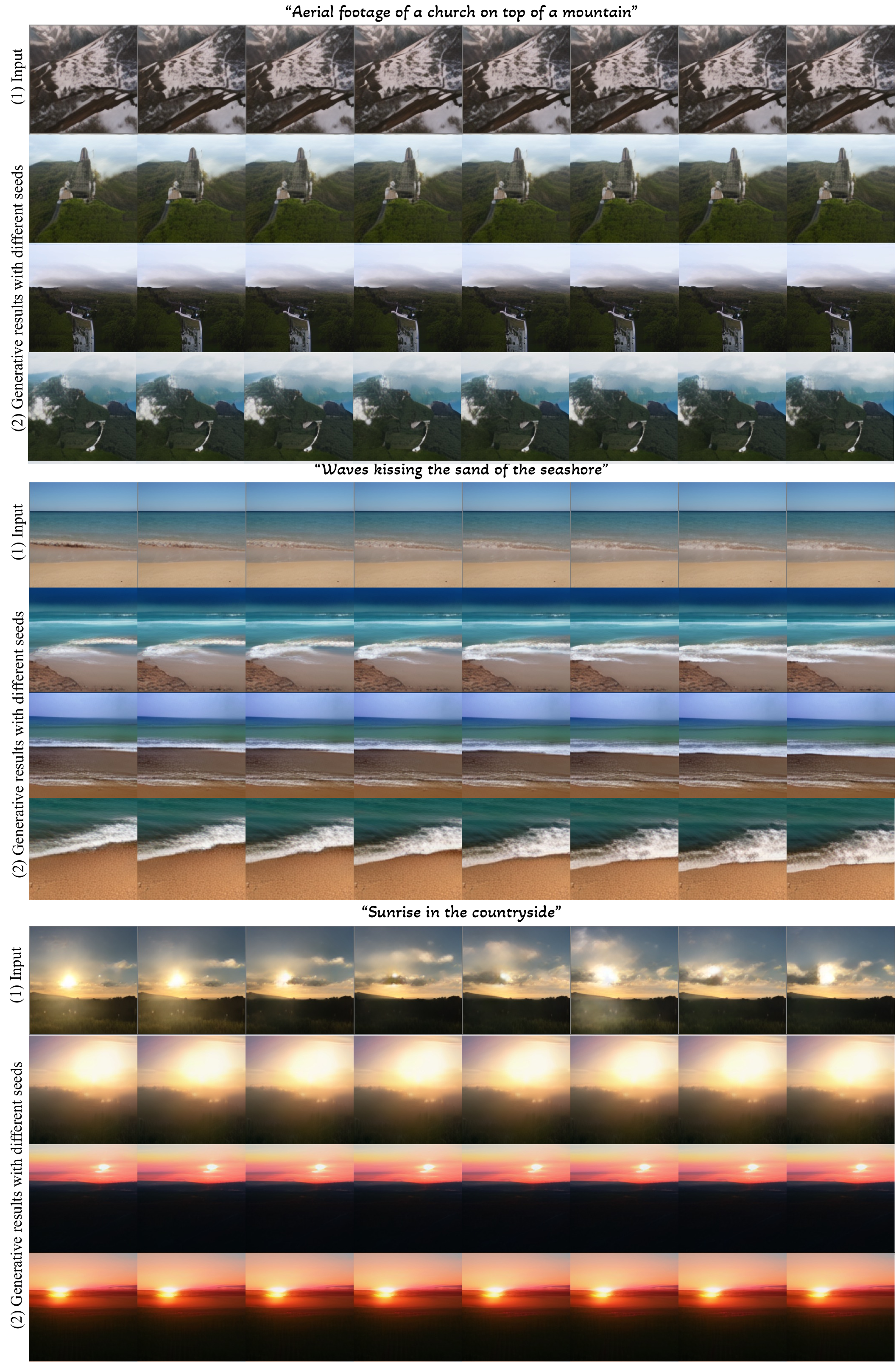}
\end{center}
\vspace{-6mm}
\caption{Generated results compared with the original training video from Pexels.}
\label{fig:seed}
\end{figure*}

The illustrative examples of this evaluation are provided in the Fig.~\ref{fig:seed}, which comprises two main components as follows:

(1) are reference input video from the Pexels training dataset, serving as the baseline for semantic and visual comparison;

(2) are multiple videos generated by our model using different random seeds, each conditioned on the caption associated with the reference video. Specifically, each row in component (2) corresponds to a distinct caption, clearly presenting the original training video alongside multiple generated variations.

This comparative visual analysis effectively illustrates both the fidelity to semantic intent and the degree of creative variation introduced by the model. Generated videos consistently reflect the semantic content of the captions while introducing sufficient diversity to avoid redundancy or overly constrained outputs. Furthermore, this evaluation underscores the model's proficiency in generating content that, though inspired by the training data, is distinctly original and enriched with novel, creative details. Such capabilities make the model highly suitable for a wide array of practical applications, ranging from creative media generation to automated content production.

\end{document}